\documentclass[tinyml]{acmart}
\usepackage{multicol}
\usepackage{multirow}
\usepackage{setspace} 
\usepackage[ruled,vlined]{algorithm2e}
\AtBeginDocument{%
  \providecommand\BibTeX{{%
    \normalfont B\kern-0.5em{\scshape i\kern-0.25em b}\kern-0.8em\TeX}}}

\setcopyright{rightsretained}
\copyrightyear{2022}
\acmYear{2022}

\begin{document}

\title{Toward Compact Deep Neural Networks via Energy-Aware Pruning}

\author{Seul-Ki Yeom}
\authornote{Corresponding author}
\email{skyeom@nota.ai}
\orcid{1234-5678-9012}
\author{Kyung-Hwan Shim}
\email{kyunghwan.shim@nota.ai}
\affiliation{%
  \institution{Nota AI}
  \city{Seoul}
  \country{Republic of Korea}
}

\author{Jee-Hyun Hwang}
\email{hwanggh96@yonsei.ac.kr}
\affiliation{%
  \institution{Yonsei University}
  \city{Seoul}
  \country{Republic of Korea}
}

\renewcommand{\shortauthors}{Yeom, et al.}

\begin{abstract}
Despite the remarkable performance, modern deep neural networks are inevitably accompanied by a significant amount of computational cost for learning and deployment, which may be incompatible with their usage on edge devices. Recent efforts to reduce these overheads involve pruning and decomposing the parameters of various layers without performance deterioration. Inspired by several decomposition studies, in this paper, we propose a novel energy-aware pruning method that quantifies the importance of each filter in the network using nuclear-norm (NN).
Proposed energy-aware pruning leads to state-of-the-art performance for Top-1 accuracy, FLOPs, and parameter reduction across a wide range of scenarios with multiple network architectures on CIFAR-10 and ImageNet after fine-grained classification tasks. On toy experiment, without fine-tuning, we can visually observe that NN has a minute change in decision boundaries across classes and outperforms the previous popular criteria. We achieve competitive results with 40.4/49.8\% of FLOPs and 45.9/52.9\% of parameter reduction with 94.13/94.61\% in the Top-1 accuracy with ResNet-56/110 on CIFAR-10, respectively. In addition, our observations are consistent for a variety of different pruning setting in terms of data size as well as data quality which can be emphasized in the stability of the acceleration and compression with negligible accuracy loss.
\end{abstract}

\maketitle

\section{Introduction}
\begin{figure}[t!]
	\begin{center}
		\centerline{\includegraphics[width=1\linewidth]{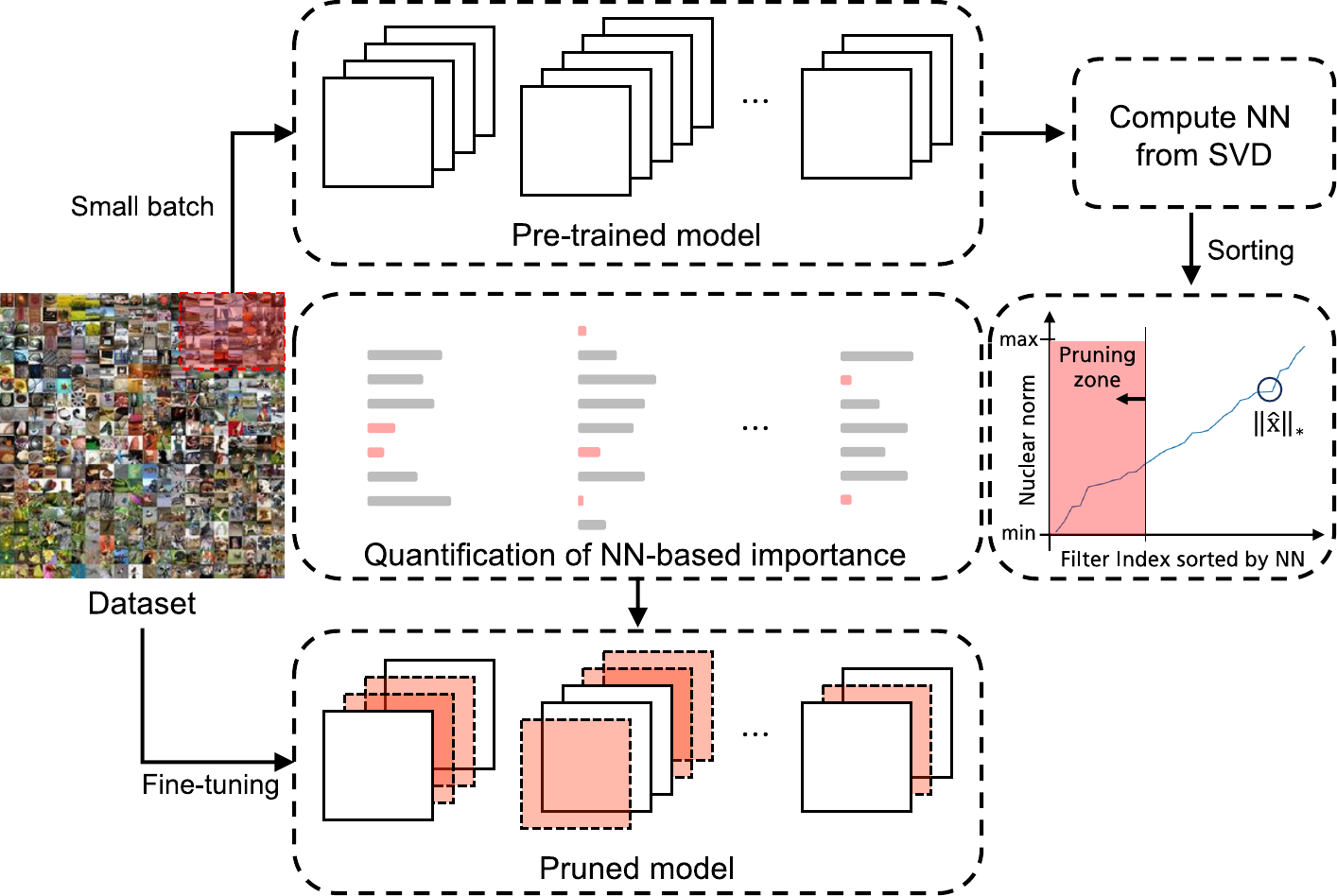}}
		\caption{Framework of the proposed method for pruning. After flattening/concatenating each filter maps for all inputs, SVD is applied to retrieve nuclear-norm values. Then the actual pruning process takes place as presented in the bottom-most column, according to the ordering calculated for each layer.}
		\label{plot:framework}
	\end{center}
\end{figure}
Deep Neural Networks (DNNs) have achieved great successes in various applications such as image classification~\cite{tan2019efficientnet}, detection~\cite{Tan_2020_CVPR}, and semantic segmentation~\cite{tao2020hierarchical}. However, these modern networks require significant computational costs and storage, making it difficult to deploy in real-time applications without the support of a high-efficiency Graphical Processing Unit (GPU). To address this issue, various network compression methods such as pruning~\cite{GM, luo2020autopruner, adaptive1, adaptive2, wang2021convolutional}, quantization~\cite{jacob2018quantization, quantization1}, low-rank approximation~\cite{CP_LRD, NIPS_LRD}, and knowledge distillation~\cite{hinton2015distilling, kd1} are constantly being developed.

Among diverse network compression strategies, network pruning has steadily grown as an indispensable tool, aiming to remove the least important subset of network units (i.e. neurons or filters) in the structured or unstructured manner. For network pruning, it is crucial to decide how to identify the ``irrelevant'' subset of the parameters meant for deletion. To address this issue, previous researches have proposed specific criteria such as Taylor approximation, gradient, weight, Layer-wise Relevance Propagation (LRP), and others to reduce complexity and computation costs in the network. Recently several studies, inspired by low-rank approximation which can efficiently reduce the rank of the corresponding matrix, have been started from the viewpoint of pruning~\cite{hrank,Li2020CVPR}. Indeed, pruning and decomposition have a close connection like two sides of the same coin from the perspective of compression~\cite{Li2020CVPR}.

The concept of the decomposition-based compression studies proposes that the network is compressed by decomposing a filter into a set of bases with singular values on a top-$k$ basis, in which singular values represent the importance of each basis~\cite{XueLG13}. In other word, we can say that decomposition allows to optimally conserve the \textit{energy}, which can be a summation of singular values~\cite{energy_define}. From the macroscopic point of view, we here believe that the energy-aware components could be used as an efficient criterion to quantify the filters in the network. 

We propose an energy-aware pruning method that measures the importance scores of the filters by using an energy-based criterion inspired by previous filter decomposition methods. More specifically, we compute nuclear-norm (NN) derived from singular values decomposition (SVD) to efficiently and intuitively quantify the filters into an energy cost. Our experimental results show that the NN-based pruning can lead the state-of-the-art performance regardless of network architectures and datasets, assuming that the more/less energy contains, the better/worse filter stands for. We prune the filters with the least energy throughout the network. A detailed description of the overall framework of our energy-aware pruning process is shown in Fig.~\ref{plot:framework}.

\begin{figure}[t!]
	\begin{center}
		\centerline{\includegraphics[width=1\linewidth]{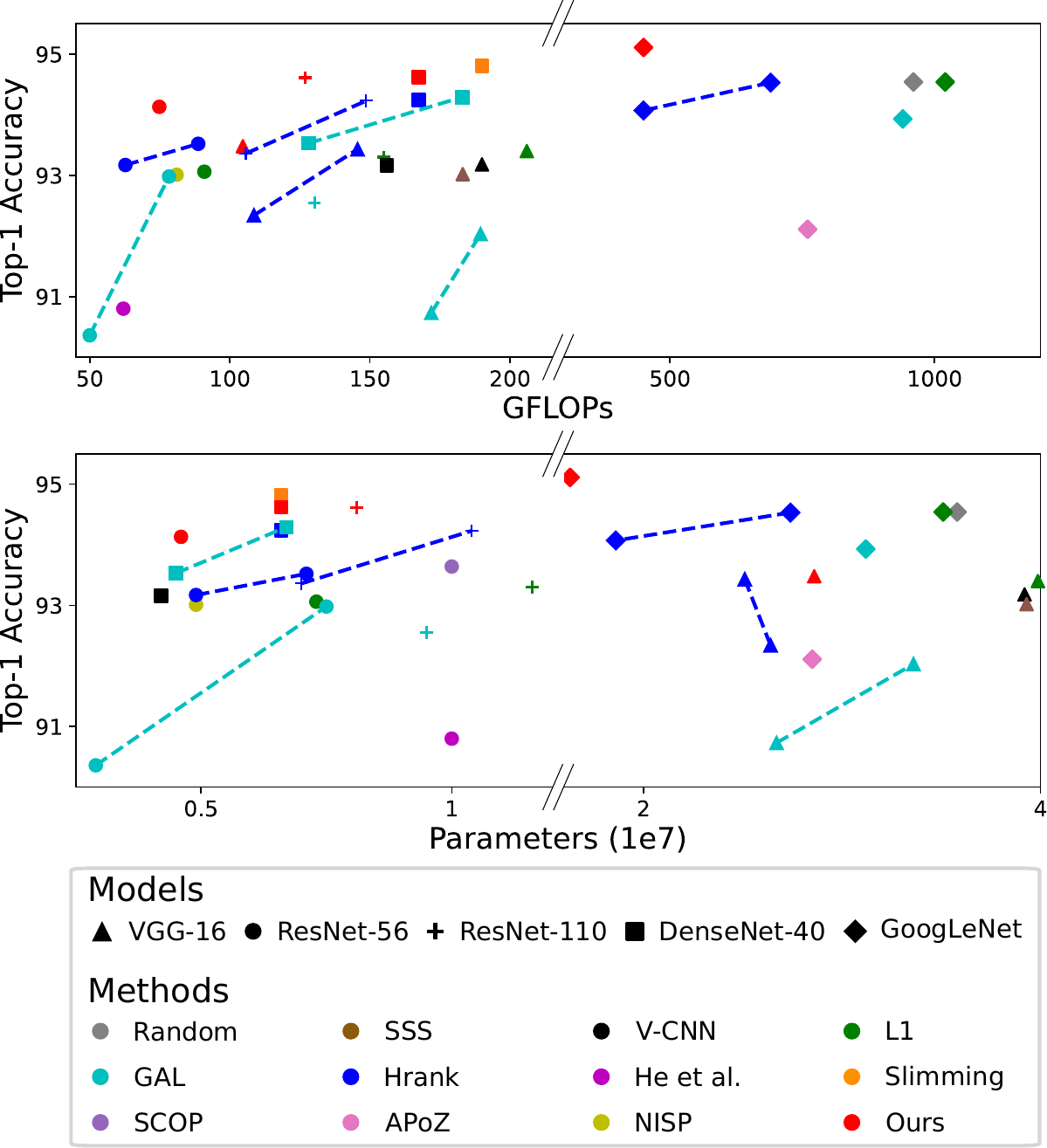}}
		\caption{Comparing between accuracy and FLOPs (top) and accuracy and total number of remained parameters (bottom) with five network architectures (VGG-16, ResNet-56, ResNet-110, DenseNet-40, and GoogLeNet) on CIFAR-10 dataset. Top-left is better performance.}
		\label{plot:result_1}
	\end{center}
\end{figure}

To summarize, our main contributions are:
\begin{itemize}
    \item We introduce a novel energy-aware pruning criterion for filter pruning which removes filters with the lowest nuclear-norm that can be quantified which lead to efficiently reducing network complexity. Results prove the efficiency and effectiveness of our proposed method through extensive experiments.
    \item Nuclear-norm based energy-aware pruning achieves state-of-the-art performances with a similar compression ratio over a variety of existing pruning approaches \cite{GM, adaptive1, SSS, hrank, GAL, liunetworkslimming, luo2017thinet, Yu00LMHGLD18, zhao2019variational} on all kinds of network architectures, as shown in Figure~\ref{plot:result_1}.
    \item Furthermore, the proposed NN-based pruning approach can lead high stability over the quality and quantity of the data, which is greatly beneficial to the practical industry aspect. This property of the proposed method is described in detail in Ablation study.
\end{itemize}
\section{Related Works}
\label{sec:related_works}
\subsection{Filter Decomposition}
Filter decomposition approaches decompose network matrices into several bases for vector spaces to estimate the informative parameters of the DNNs with low-rank approximation/factorization, thus reducing computation cost of the network~\cite{li2019learning} such as SVD~\cite{NIPS_LRD}, CP decomposition~\cite{CP_LRD}, Tucker decomposition~\cite{KimPYCYS15}, and others \cite{jaderberg2014speeding} approximate convolutional operations by representing the weight matrix as smaller bases set of 2D separable filters without changing the original number of filters. In ~\cite{filter_distillation}, Principal Component Analysis (PCA) was applied on max-pooled and flattened feature maps, to compute the amount of information to be preserved in each layer among all layers, enabling integration with each other. 


\subsection{Filter Pruning}
Network filter pruning removes redundant or non-informative filters which are less informative for performance from the given model at once (one-shot pruning) or iteratively (iterative pruning). Most network filter pruning techniques make filters sparse by removing connections and adopting an appropriate criterion for discriminating whether it is crucial or not. It is a critical point to decide how to quantify the importance of the filters in the current state of the model for deletion. In previous studies, pruning criteria have been typically proposed based on the magnitude of 1) mostly weights with $l_1$~/~$l_2$-norm~\cite{NIPS2015_5784, l1norm}, 2) gradients~\cite{sun2017meprop}, 3) Taylor expansion~/~2$^{nd}$ partial derivative (a.k.a. Hessian matrix)~\cite{NIPS1989_250, taylor}, 4) Layer-wise relevance propagation (LRP)~\cite{yeompr2021}, and 4) other criteria~\cite{Yu00LMHGLD18, liunetworkslimming}. For more detail in magnitude-based pruning, please refer to~\cite{yeompr2021}.

\subsection{Pruning by decomposition}
Concurrently with our work, there is a growing interest in compressing DNNs motivated by decomposition in terms of pruning as well as fusion approach~\cite{li2020penni, Li2020CVPR, TRP2020, hrank}. Due to the close connection between two different compression methods, those works demonstrate that the decomposition-based approach can enhance the performance for efficiently compressing the model even at the filter level. \cite{li2020penni} proposes a hardware-friendly CNN model compression framework, PENNI, which applies filter decomposition to perform a small number of basis kernel sharing and adaptive bases and coefficients with sparse constraints. \cite{Li2020CVPR} proposes a unified framework that allows combining the pruning and the decomposition approaches simultaneously using group sparsity. \cite{TRP2020} proposed Trained Ranking Pruning (TRP) which integrates low-rank approximation and regularization into the training process. To constrain the model into low-rank space, they adopt a stochastic sub-gradient descent optimized nuclear-norm regularization which is utilized as a different purpose from our proposed method. Similarly to our work, \cite{hrank} proposes a high rank-based pruning method as a criterion by computing the full rank of each feature map from SVD layer-by-layer, which leads to inconsistent rank order regardless of batch size.


\section{Method}
\label{sec:method}

\subsection{Preliminaries}




From a pre-trained CNN model, we first define trainable parameters, weights as $\mathbf{W}_l=\{ \mathbf{w}_l^{1}, \mathbf{w}_l^{2}, \dots, \mathbf{w}_l^{c_{l}} \} \in \mathbb{R}^{c_{l} \times c_{l-1} \times k \times k}$, where $c_{l-1}$ and $c_{l}$ denote the number of the input and output channels and $k$ is the height/width of the squared kernel at $l$th convolutional layer. Please note that for the sake of simplicity, we omit biases term here.

Pruning has been started with a pre-trained full-size network $f(\mathcal{X};\mathbf{W})$ which is overparameterized throughout the network. For DNN, our original objective function is to minimize our loss given dataset and parameters $\mathbf{W}$.

\begin{equation}
    \label{eq:obj_func}
    \min_{\mathbf{W}} \mathcal{L} (\mathcal{Y}, f(\mathcal{X};\mathbf{W})) \\
\end{equation}
where $\mathcal{X} \in \{ \mathbf{x}_0, \mathbf{x}_1, \dots, \mathbf{x}_N \}$ and $\mathcal{Y} \in \{ \mathbf{y}_0, \mathbf{y}_1, \dots, \mathbf{y}_N \}$ represent a set of paired training inputs and its labels, respectively. $N$ denotes the total number of batches.



\begin{algorithm}[t]
\SetAlgoLined
\KwIn{Pre-trained model $f$, training data $\mathcal{X}$, pruning ratio $r$, and pruning threshold $t$}
\KwOut{Pruned model $f'$}
\Begin{
    \While{$t$ not reached}{
        // Assess network substructure importance\\
        \For{\texttt{BN layer} \textbf{in} $f$}{
            \For{\texttt{channels} in \texttt{BN layer}}{
                $\rhd$ compute equation ~\ref{eq:svd_1} and ~\ref{eq:svd_3} \\
            }
        }
        // Identify and remove least important filters in groups of $r$ \\
        $\rhd$ Remove $r$ \texttt{channels} with the lowest $\vert \vert \mathbf{\hat{x}} \vert \vert _ \ast$ from $f$ \\
        $\rhd$ Remove its corresponding connections of each removed \texttt{channel} \\
        \If{desired}{
            // Optional fine-tuning to recover performance \\
            $\rhd$ fine-tune $f'$ on $\mathcal{X}$ \\
        }
    }
    \Return{pruned model $f'$}
}
\caption{Energy-Aware Pruning}
\label{alg:pruning}
\end{algorithm}

\begin{figure*}[ht!]
	\begin{center}
		\centerline{\includegraphics[width=0.6\linewidth]{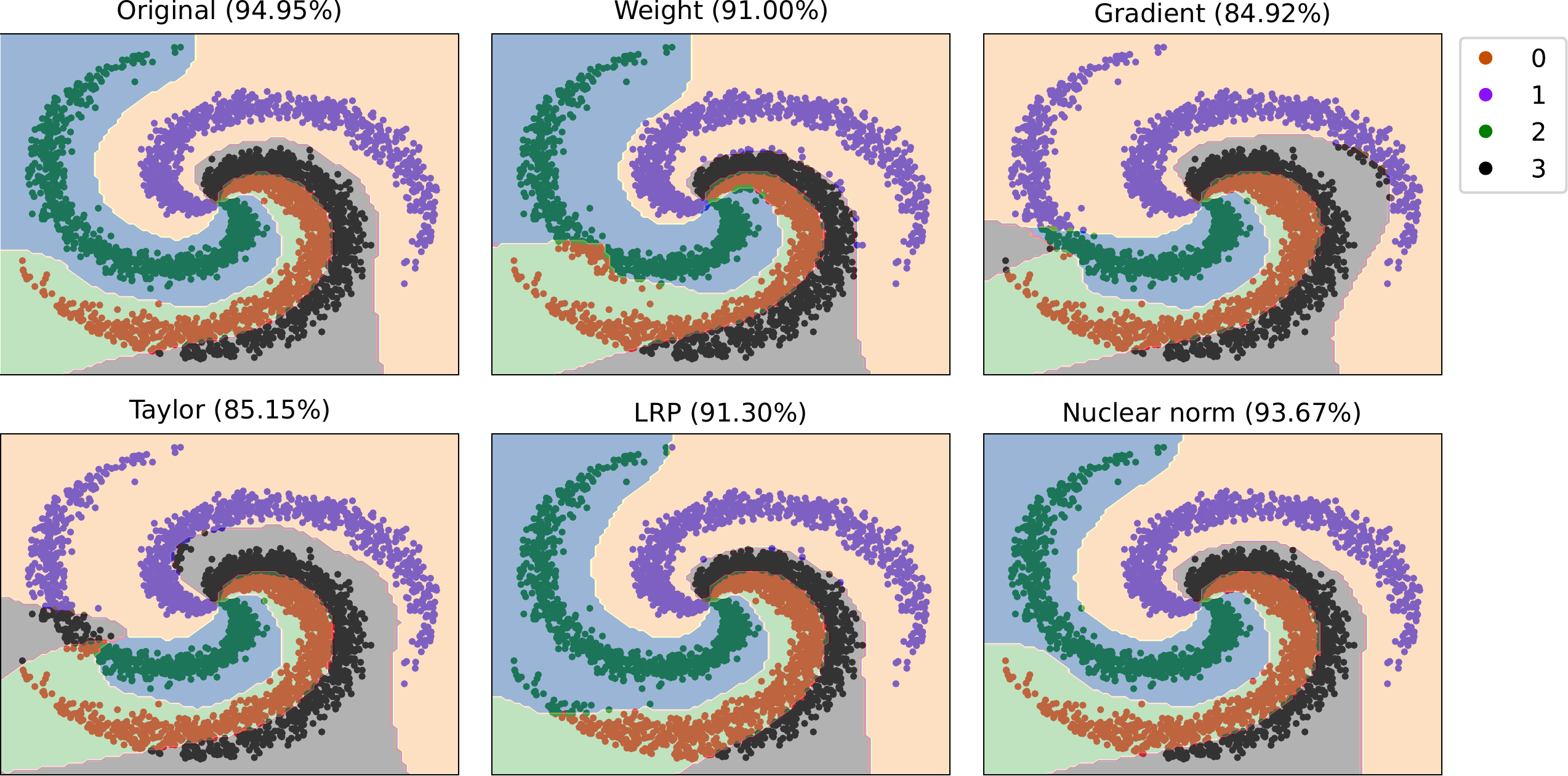}}
		\caption{Qualitative Comparison of the impact of the pruning criteria -- Original model, Weight, Gradient, Taylor, LRP, and Nuclear-norm (from left top to right bottom) --on the decision function with toy dataset (k = 4). Scores in bracket indicate accuracy after pruning 33.3\% filters of the original model followed by no fine-tuning.
		}
		\label{plot:toy_experiment}
	\end{center}
\end{figure*}

\subsection{Energy-based Filter Pruning Approach} \label{EBFPA}
Given the pre-trained model~$f(\mathcal{X};\mathbf{W})$, we define a $\mathcal{R}$ function by adopting an energy-aware pruning criterion. Our hypothesis is that the more energy a filter has, the larger amount of information it contains. In other words, we could define a regularization function that can minimize the difference between the energies from the pre-trained model and the pruned model. Therefore, in terms of energy efficiency, $\mathcal{R}$ can be defined as 
\begin{equation}
    \label{eq:method3}
    \mathcal{R}(\mathbf{W}) = \lvert \mathbf{E} (\mathcal{X};\mathbf{W}) - \mathbf{E} (\mathcal{X};\mathbf{W}') \rvert \\
\end{equation}
where $\mathbf{E}(\cdot) = \{ \mathbf{e}_{1}, \mathbf{e}_{2}, \dots, \mathbf{e}_{l} \}$ indicate total amount of energy in the network. And each $\mathbf{e}_l$ denotes the amount of energy at layer $l$ and is computed on the corresponding feature map using our criterion which will be discussed thoroughly afterward. Additionally, we introduce a binary pruning mask $\mathbf{M} \in \{ 0, 1 \} ^{c_l}$ which determines if a filter is remained or pruned during feed-forward propagation such that when $\mathbf{M}$ is vectorized: $\mathbf{W}' = \mathbf{W}\odot\mathbf{M}$, where is an element-wise multiplication between $\mathbf{W}$ and $\mathbf{M}$. And here, we assume that each $\mathbf{e}_l$ can be approximated by $\mathbf{e}_l \approx \vert \vert \mathbf{w}_l \vert \vert _ \ast $ computed by decomposition approach. Here, we adopt the decomposition approach, SVD, to quantify filter-wise energy consumption. SVD is the basis for many related techniques in dimensionality reduction used to obtain reduced-order models (ROMs). For pruning, SVD helps find the best $k$-dimensional perpendicular subspace with respect to the dataset in each point. Especially, singular values play an important role in algebraic complexity theory. That is, the singular value represents the energy of each rank-one matrix. Singular values represent the importance of its associated rank-one matrix.

Previous research showed that filter pruning and decomposition are highly related from the viewpoint of compact tensor approximation~\cite{Li2020CVPR}. There is a hinge point between both strategies in investigating a compact approximation of the tensors despite the usage of different operations in a variety of application scenarios. 
Decomposition is done to quantify the energy on the output channels in batch normalization (BN) layers. Additional to the efficient trade-off of channel-level sparsity, BN provides normalized values of the internal activation using mini-batch statistics to any scale~\cite{liunetworkslimming}. This process is achieved by applying 3D channel outputs $\mathbf{\hat{x}}_l$ at $l^{th}$ BN layer. The superscript $l$ which indicates the number of layer is omitted for readability. Based on $\mathbf{\hat{x}}$, we first reshape the original 3D tensor into a 2D tensor $\mathbf{\hat{x}}$ along with the channel.

From the SVD, a channel output at $l^{th}$ layer can be decomposed as follow,
\begin{equation}
    \label{eq:svd_1}
    \mathbf{\hat{x}} = U S V^T = \sum_{i=1}^{N} \sigma_i u_i v_i^T
\end{equation} 
where $U$ and $V$ denote the left and right singular vector matrix respectively and $S$ indicates the diagonal matrix of singular values $\sigma_n$ where $S = diag(\sigma_1, \sigma_2, \dots, \sigma_N)$.

\begin{equation}
    \label{eq:svd_3}
    \vert \vert \mathbf{\hat{x}} \vert \vert _ \ast = \sum_{i=1}^{N} \sigma_i
\end{equation}

$\vert \vert \mathbf{\hat{x}} \vert \vert _ \ast$ denotes \textit{nuclear-norm}, the sum of the singular values which can represent the energy of the model~\cite{DBLP:journals/corr/Sadek12}. Here, based on our hypothesis, a useful rule of thumb for efficient filter pruning is to optimally preserve the energy throughout the network. In this respect, based on equation~\ref{eq:svd_3}, we can not only evaluate the distribution of the feature spaces but also estimate the contribution of the feature spaces simultaneously, which can be applicable for a pruning criterion. Additionally, it provides necessary and sufficient conditions for rank consistency while minimizing the loss of the model ~\cite{Bach08}. For this reason, it leads to achieving consistent results regardless of data quality as well as data quantity. The procedure based on the pruning method is outlined in Algorithm~\ref{alg:pruning}.

\section{Experiments}
\label{sec:experiments}
\subsection{Experimental Setup}
\subsubsection{Models and Datasets}
We demonstrate the effectiveness of the proposed energy-aware pruning with nuclear-norm on four types of pre-trained feed-forward deep neural network architectures from various perspective comparison studies: 1) VGG-16~\cite{vgg}), ResNet-56, ResNet-110~\cite{resNet}, GoogLeNet~\cite{googlenet}, DenseNet-40~\cite{densenet} for CIFAR-10~\cite{cifar10}, and ResNet-50 on ImageNet~\cite{deng2009imagenet}. The resolution of each image is 32$\times$32 (CIFAR-10) and 224$\times$224 (ImageNet) pixels, respectively. 

\subsubsection{Implementation Details}
We conduct all pruning experiments on \textit{Pytorch 1.6} under \textit{Intel(R) Xeon(R) Silver 4210R CPU 2.40GHz} and \textit{NVIDIA RTX 2080Ti with 12GB} for GPU processing. After one-shot pruning, we adopt the Stochastic Gradient Descent (SGD) algorithm as an optimization function. For both the CIFAR-10 and ImageNet, over-parameterized models are pruned at a time and fine-tuned by using 200 epochs with early stopping with 0.01 initial learning rate, scheduled by using cosine scheduler. Cross entropy is selected as a loss function. And the momentum and the weight decay factor are 0.9 and $5\times10^{-4}$, respectively. And we set the fine-tuning batch size of 128. For pruning, we adopt the built-in function \textit{torch.nn.utils.prune} in \textit{PyTorch} throughout the experiments.

\subsubsection{Evaluation Metrics}
For a fair competition, we measure Top-1 accuracy (CIFAR-10 and ImageNet) and Top-5 accuracy (ImageNet only) of the pruned network as baselines. Also, we computed the Floating point operations (FLOPs) as well as the total remaining number of parameters (params) to precisely compare the efficiency of the proposed criterion in terms of computational efficiency.

\subsection{Results on Toy Experiment}
First, we start by comparing the properties and effectiveness of the several pruning criteria on the toy dataset. In addition to our proposed criterion (i.e. nuclear-norm), we also evaluate against existing importance-based pruning criteria on the toy dataset: weight, gradient, Taylor expansion, and layer-wise relevance propagation (LRP). For more experimental settings on toy experiments, please find Supplementary~\ref{sup:toy_example}.

Figure~\ref{plot:toy_experiment} shows the data distributions of the generated multi-class toy datasets to see the qualitative impact on the models' decision boundary when removing 1000 neurons with considered pruning criteria. This demonstrates how the toy models' decision boundaries change under influence of pruning with all five criteria. We can observe that both the Taylor and gradient measures degrade the model significantly whereas weight and LRP preserve the decision boundary from the pruned models reasonably except for the area where classify between 1) class No. 0 (brown) and class No. 2 (green) and 2) between class No. 0 and class No. 3 (black). On the other hand, we can clearly see that in contrast to the other property importance-based pruning criteria, nuclear-norm significantly classify multi-classes even after the pruning process, thus allowing to safely remove the unimportant (w.r.t. classification) elements. As we can see in Figure~\ref{plot:toy_experiment}, NN-based pruning results in only minimal change in the decision boundary, compared to the other criteria. Furthermore, nuclear-norm can successfully preserve original accuracy of 94.95\% up to 93.67\% whereas 91.00\% of weight, 84.92\% of gradient, 85.15\% of Taylor expansion, and 91.30\% of LRP.

\begin{table}[!ht]
    \caption{Pruning results of five network architectures on CIFAR-10. Scores in brackets of ``FLOPs'' and ``Params'' denote the compression ratio of FLOPs and parameters in the compressed models.}
    \scalebox{0.80}{
        \begin{tabular}{ccccc}
            \specialrule{1pt}{1pt}{1pt}
            \hline
            Criterion & Pruned & Gap & FLOPs ($\downarrow$\%) & Params ($\downarrow$\%) \\ \hline
            \multicolumn{5}{c}{VGG-16-BN}                                                                                                \\
            L1~\cite{l1norm}                         & 93.40             & 0.15            & 206.00M (34.3)              & 5.40M (64.0)                 \\
            Variational CNN~\cite{zhao2019variational}            & 93.18            & -0.07           & 190.00M (39.4)              & 3.92M (73.3)                 \\
            SSS~\cite{SSS}                        & 93.02            & -0.94           & 183.13M (41.6)              & 3.93M (73.8)                 \\
            GAL-0.05~\cite{GAL}                   & 92.03            & -1.93           & 189.49M (39.6)              & 3.36M (77.6)                 \\
            GAL-0.1~\cite{GAL}                    & 90.73            & -3.23           & 171.89M (45.2)              & 2.67M (82.2)                 \\
            HRank-53~\cite{hrank}                      & 93.43            & -0.53           & 145.61M (53.5)              & 2.51M (82.9)                 \\
            HRank-65~\cite{hrank}                      & 92.34            & -1.62           & 108.61M (65.3)              & 2.64M (82.1)                 \\
            \textbf{Propose method}             & \textbf{93.48}            & \textbf{-0.48}           & \textbf{104.67M (66.6)}              & \textbf{2.86M (80.9)}                 \\ \hline
            \multicolumn{5}{c}{ResNet-56}                                                                                                \\
            L1~\cite{l1norm}                         & 93.06            & 0.02            & 90.90M (27.6)               & 0.73M (14.1)                 \\
            NISP~\cite{Yu00LMHGLD18}                       & 93.01            & -0.25           & 81.00M (35.5)               & 0.49M (42.4)                 \\
            GAL-0.6~\cite{GAL}                    & 92.98            & -0.28            & 78.30M (37.6)               & 0.75M (11.8)                 \\
            GAL-0.8~\cite{GAL}                    & 90.36            & -2.90           & 49.99M (60.2)               & 0.29M (65.9)                 \\
            He~\textit{et al.}~\cite{adaptive1}& 90.80& -2.00              & 62.00M (50.6)               & -                          \\
            HRank-29~\cite{hrank}                      & 93.52            & 0.26            & 88.72M (29.3)               & 0.71M (16.8)                 \\
            HRank-50~\cite{hrank}                      & 93.17            & -0.09           & 62.72M (50.0)               & 0.49M (42.4)                 \\
            SCOP~\cite{SCOP}                       & 93.64            & -0.06           & 54.83M (56.3)                  & 0.37M (56.0)                   \\
            \textbf{Propose method}             & \textbf{94.13}            & \textbf{0.87}            & \textbf{74.83M (40.4)}               & \textbf{0.46M (45.9)}                 \\ \hline
            \multicolumn{5}{c}{ResNet-110}                                                                                               \\
            L1~\cite{l1norm}                         & 93.30             & -0.20            & 155.00M (38.7)              & 1.16M (32.6)                 \\
            GAL-0.5~\cite{GAL}                    & 92.55            & -0.95           & 130.20M (48.5)              & 0.95M (44.8)                 \\
            HRank-41~\cite{hrank}                      & 94.23            & 0.73            & 148.70M (41.2)              & 1.04M (39.4)                 \\
            HRank-58~\cite{hrank}                      & 93.36            & -0.14           & 105.70M (58.2)              & 0.70M (59.2)                 \\
            \textbf{Propose method}             & \textbf{94.61}            & \textbf{1.11}            & \textbf{126.96M (49.8)}              & \textbf{0.81M (52.9)}                 \\ \hline
            \multicolumn{5}{c}{GoogLeNet}                                                                                                \\
            Random& 94.54            & -0.51           & 0.96B (36.8)                & 3.58M (41.8)                 \\
            L1~\cite{l1norm}                           & 94.54            & -0.51           & 1.02B (32.9)                & 3.51M (42.9)                 \\
            APoZ~\cite{apoz}                       & 92.11            & -2.94           & 0.76B (50.0)                & 2.85M (53.7)                 \\
            GAL-0.5~\cite{GAL}                    & 93.93            & -1.12           & 0.94B (38.2)                & 3.12M (49.3)                 \\
            HRank-54~\cite{hrank}                      & 94.53            & -0.52           & 0.69B (54.9)                & 2.74M (55.4)                 \\
            HRank-70~\cite{hrank}                      & 94.07            & -0.98           & 0.45B (70.4)                & 1.86M (69.8)                 \\
            \textbf{Propose method}             & \textbf{95.11}            & \textbf{0.06}            & \textbf{0.45B (70.4)}                & \textbf{1.63M (73.5)}                 \\ \hline
            \multicolumn{5}{c}{DenseNet-40}                                                                                              \\
            Network Slimming~\cite{liunetworkslimming}           & 94.81            & -0.92            & 190.00M (32.8)                 & 0.66M (36.5)                 \\
            GAL-0.01~\cite{GAL}& 94.29            & -0.52            & 182.92M (35.3)              & 0.67M (35.6)                 \\
            GAL-0.05~\cite{GAL}& 93.53             & -1.28           & 128.11M (54.7)              & 0.45M (56.7)                 \\
            Variational CNN~\cite{zhao2019variational}            & 93.16            & -0.95           & 156.00M (44.8)              & 0.42M (59.7)                 \\
            HRank-40~\cite{hrank}                      & 94.24            & -0.57           & 167.41M (40.8)              & 0.66M (36.5)                 \\
            \textbf{Propose method}             & \textbf{94.62}            & \textbf{-0.19}           & \textbf{167.41M (40.8)}              & \textbf{0.66M (36.5)}                 \\ \hline
        \end{tabular}
    }
    \label{tab:cifar10_table}
\end{table}

\begin{table}[ht]
    \begin{center}
        \caption{Pruning results on  ResNet-50 with ImageNet. Scores in brackets of ``FLOPs'' and ``Params'' denote the compression ratio of FLOPs and parameters in the compressed models.}
    \scalebox{0.72}{
        \begin{tabular}{ccccccc}
            \specialrule{1pt}{1pt}{1pt}
                \multicolumn{7}{c}{ResNet-50}                                                                                                                                                                \\ \hline
                \multirow{2}{*}{Criterion}& \multicolumn{2}{c}{Top-1 Acc (\%)} & \multicolumn{2}{c}{Top-5 Acc (\%)} & \multirow{2}{*}{FLOPs ($\downarrow$\%)} & \multirow{2}{*}{Params ($\downarrow$\%)} \\
                & Pruned& Gap& Pruned& Gap&&                                    \\ \hline
                He \textit{et al.}~\cite{adaptive1}& 72.30             & -3.85           & 90.80             & -1.40& 2.73B (33.25)& -\\
                ThiNet-50~\cite{luo2017thinet}& 72.04            & -0.84           & 90.67            & -0.47           & 2.58B (36.8)                         & 16.90M (33.72)\\
                SSS-26~\cite{SSS}& 71.82& -4.33           & 90.79            & -2.08           & 2.33B (43.0)                       & 15.60M (38.8)                      \\
                SSS-32~\cite{SSS}& 74.18& -1.97           & 91.91            & -0.96           & 2.82B (31.0)                       & 18.60M (27.0)                      \\
                GAL-0.5~\cite{GAL}& 71.95& -4.20            & 90.94            & -1.93           & 2.33B (43.0)                       & 21.20M (16.8)                      \\
                GAL-0.5-joint~\cite{GAL}& 71.80             & -4.35           & 90.82            & -2.05           & 1.84B (55.0)                       & 19.31M (24.2)                      \\
                GAL-1~\cite{GAL}& 69.88& -6.27           & 89.75            & -3.12           & 1.58B (61.3)                       & 14.67M (42.4)                      \\
                GAL-1-joint~\cite{GAL}& 69.31            & -6.84           & 89.12            & -3.75           & 1.11B (72.8)                       & 10.21M (59.9)                      \\
                GDP-0.5~\cite{globalPruningIJCAI}& 69.58& -6.57           & 90.14            & -2.73           & 1.57B (61.6)                       & -                               \\
                SFP~\cite{l2norm}& 74.61            & -1.54           & 92.06            & -0.81           & 2.38B (41.8)                       & -                                 \\
                AutoPruner~\cite{luo2020autopruner}& 74.76            & -1.39           & 92.15            & -0.72           & 2.09B (48.7)                       & -\\
                FPGM~\cite{GM}& 75.59            & -0.56           & 92.27            & -0.60            & 2.55B (37.5)                       & 14.74 (42.2)                       \\
                Taylor~\cite{taylor_result}& 74.50             & -1.68           & -              & -             & 2.26B (44.5)& 14.05M (44.9)\\
                RRBP~\cite{zhou2019accelerate}& 73.00               & -3.10            & 91.00& -1.90& -& 11.60M (54.5)\\ 
                GDP-0.6~\cite{globalPruningIJCAI}& 71.19& -4.96           & 90.71            & -2.16           & 1.88B (54.0)                       & -                                \\
                HRank-74~\cite{hrank}& 74.98            & -1.17           & 92.33            & -0.54           & 2.30B (43.7)                       & 16.15M (36.6)                      \\
                HRank-71~\cite{hrank}& 71.98            & -4.17           & 91.01            & -1.86           & 1.55B (62.1)                       & 13.77M (46.0)                      \\
                HRank-69~\cite{hrank}& 69.10             & -7.05           & 89.58            & -3.29           & 0.98B (76.0)                       & 8.27M (67.5)                       \\
                SCOP~\cite{SCOP}& 75.26            & -0.89           & 92.53            & -0.34           & 1.85B (54.6)                       & 12.29M (51.8)                      \\
                SRR-GR~\cite{wang2021convolutional}& 75.11\% &-1.02 &92.35\% & -0.51&1.83B (55.1\%)& - \\\hline
                \multirow{2}{*}{\textbf{Propose method}} & \textbf{75.25}  & \textbf{-0.89} & \textbf{92.49}  & \textbf{-0.37} & \textbf{1.52B (62.8)}& \textbf{11.05M (56.7)}\\
                                                         & \textbf{72.28}   & \textbf{-3.87}  & \textbf{90.93}  & \textbf{-1.93} & \textbf{0.95B (76.7)}& \textbf{8.02M (68.6)} \\
            \specialrule{1pt}{1pt}{1pt}
        \end{tabular}
        \label{tab:imagenet_table}
    }
    \end{center}
\end{table}
\subsection{Results on CIFAR-10}
To prove the expandability of the proposed nuclear-norm based pruning approaches on the various deep learning-related modules, such as residual connection or inception module, we compress several popular DNNs, including VGG-16, ResNet-56/110, GoogLeNet, and DenseNet-40. Due to the different original performance of each literature, we then report the performance gap between their original model and the pruned model. All results are presented in Table~\ref{tab:cifar10_table} on the CIFAR-10 dataset.

\subsubsection*{\textbf{VGG-16}} We first test the basic DNN architecture, VGG-16, which is commonly used as a standard architecture. It can verify the efficiency of the proposed pruning method on the consecutive convolutional block. For a fair comparison study, we adopt several conventional importance-based methods -- L1, HRank, SSS, Variational CNN \textit{et al}., and GAL -- in this experiment. We reached initial Top-1 accuracy of 93.96\% with 313.73 million FLOPs and 14.98 million parameters. VGG-16 consists of 13 convolutional blocks with 4224 convolutional filters and 3 fully-connected layers. In terms of complexity, VGG-16 with batch normalization contains 313.73 million FLOPs and 14.98 million parameters initially.

The proposed nuclear-norm based pruning method outperforms previous conventional pruning approaches, especially on the performance and the FLOPs as well as parameter reduction. Most of the conventional pruning approaches could compress more than 70\% of the parameters, while they could not accelerate the VGG-16 model effectively. On the other hand, the proposed method could yield a highly accelerated model but with a tiny performance drop. To be more specific, GAL accelerates the baseline model by 45.2\% and 39.6\% while it compresses 82.2\% and 77.6\% of the model with 90.73\% and 92.03\% of the performance. However, the proposed method yields the pruned model with 66.6\% reduced FLOPs (104.67M) and 80.9\% reduced parameters (2.86M) with only 0.48\% of accuracy drop from scratch, which outperforms in all of the aspects (performance, acceleration, and compression). Compared to the recent importance-based method, HRank, which also uses the rank property for pruning, the proposed method achieves the competitive performance acceleration(93.48\% \textit{vs.} 92.34\% and 104.67M \textit{vs.} 108.61M) but with a similar compress ratio.

\subsubsection*{\textbf{ResNet-56/110}} The residual connection of the ResNet is consists of an element-wise add layer, requiring the same input shape. For this reason, pruning on ResNet needs to be carefully managed compared to pruning plain networks. To equalize those inputs of the element-wise add operation of the ResNet, we prune common indices of the connected convolutional layer. By using the nuclear-norm based pruning method and the above pruning strategy, we could yield a faster and smaller model than the other approaches.

Initial Top-1 accuracies of ResNet-56 / 110 are 93.26 / 93.50\% with 125.49 / 252.89 million of FLOPs and 0.85 / 1.72 million of parameters, respectively. 
The compressed model with the proposed pruned approaches achieves 0.87\% higher performance, while it assures a similar compression and acceleration rate with conventional approaches (40.4\% of FLOPs and 45.9\% of parameters). Most of the conventional pruning approaches could not exceed the original performance except HRank (93.52\% of Top-1 accuracy), which only accelerate and compress 29.3\% of FLOPs and 16.8\% of parameters.

Furthermore, the compressed ResNet-110 also outperforms the baseline model by 1.11\% with 40.8\% of acceleration rate and 52.9\% of compression rate. Similar to ResNet-56, the NN based pruning method achieves the highest performance on ResNet-110 with a similar acceleration and compression ratio. On the other hand, the conventional pruning approaches yield around 92.55\% - 94.23\% of Top-1 accuracies while the pruned model contains around up to 0.70 - 1.16 million of compressed parameters and 105.70 - 155 million of accelerated FLOPs. Similar to the compressed model of the proposed method, HRank also outperforms the baseline accuracy, but with the larger and slower model compared to our method. In conclusion, the compressed model of the proposed method outperforms the baseline of both ResNet-56/110, which has the potential to be compressed or accelerated more without performance deterioration.
\subsubsection*{\textbf{GoogLeNet}} Unlike the residual connection, the input kernel size of the concatenation module does not have to be equivalent, therefore, coping with the inception module is relatively straightforward. We initially achieved Top-1 accuracy of 95.05\%, 1.52 billion FLOPs, and 6.15 million parameters. The proposed nuclear-norm based method greatly reduces the model complexity (70.4\% of FLOPs and 73.5\% of parameters) while it outperforms the baseline model (95.11\%~\textit{vs.} 95.05\%). Proposed pruning approach could yield the highest performance (95.11\%) with the most limited number of parameters (73.5\%). HRank reaches the performance of 94.07\%, while it accelerates around 70.4\%, but the proposed method returns 1.04\% higher performance and prunes an additional 0.23M of the parameters. The performance and the complexity of the nuclear-norm based pruning method indicate that the GoogLeNet can be compressed and accelerated more with tolerable performance drop. It demonstrates its stability to compress and accelerate the inception module without performance degradation.
\subsubsection*{\textbf{DenseNet-40}} The original model contains 40 layers with a growth rate of 12, it achieves 94.81\% on the CIFAR-10 dataset with 282.00M of FLOPs and 1.04M of parameters. The channel-wise concatenation module of the DenseNet-40 is also treated similarly to the inception module of GoogLeNet. We followed the global pruning ratio of HRank. As a result, the proposed method could outperform by 0.38\% with the same amounts of FLOPs and parameters. The compressed model could not exceed the performance of Network slimming, however, the FLOP compression rates of the proposed model could be accelerated by 22.59M.
\subsection{Results on ImageNet}
We also test the performance with our proposed criterion on ImageNet with a popular DNN, ResNet-50. Comparison of pruning ResNet-50 on ImageNet by the proposed method and other existing methods presented can be seen in the Table~\ref{tab:imagenet_table} where we report Top-1 and Top-5 accuracies, as well as FLOPs and parameters reduction. The initial performance of ResNet-50 on ImageNet is 76.15\% and 92.87\% of Top-1 and Top-5 accuracies with 4.09 billion FLOPs and 25.50 million parameters. Compare with other existing pruning methods, it is observed that our proposed method achieves better performance in all aspects. By pruning 62.8\% of FLOPs and 56.7\% of parameters from original ResNet-50 we only lose 0.89\% and 0.37\% in Top-1 and Top-5 accuracies while compressing 2.69$\times$ of FLOPs and 2.30$\times$ of parameters at the same time. When compressing the model aggressively, we could achieve 72.28\% and 90.93\% of Top-1 and Top-5 accuracies while reducing 76.7\% of FLOPs and 68.6\% of parameters which still represent a reasonable result. 
\subsection{Ablation study}
We further conduct two additional ablation studies in the perspectives of the data quality and quantity to see whether our proposed method also yields stable performance regardless of two properties for the practical industry issue. These would be the critical points when you encounter 1) lack of data, 2) dataset with overconfidence or uncertainty for the efficient pruning. We test on two more scenarios with modern neural network architectures to see the effect of rank consistency. Please find Supplementary~\ref{sup:result_quantuty} for the results in data quantity.

\section{Conclusion}
\label{sec:conclusion}
Behind the remarkable growth of modern deep neural networks, millions of trainable parameters remain an unsolved problem. After training, the extremely high cost for inference time remains one of the main issues in the entire machine learning applications. In this paper, we propose a novel energy-aware criterion that prunes filters to reduce network complexity using nuclear-norm motivated by decomposition/approximation-based approaches. Empirically, we demonstrated that the proposed criterion outperforms prior works on a variety of DNN architectures in terms of accuracy, FLOPs as well as the number of compressed parameters. Furthermore, it can be applicable for 
the specific scenarios which limit on data quantity (e.g. pruning after transfer learning and few-shot learning which small amount of dataset are required) and data quality (e.g. consisting of over-confident/uncertainty data)

For further research, more experiments can be done on 1) a unified framework in which pruning is followed by decomposition of pre-trained models to simultaneously achieve a small drop in accuracy (by pruning) and reduced FLOPs and parameters for the fast inference time (by decomposition) 2) eXplainable Artificial Intelligence (XAI) approach using our proposed method.

\begin{acks}
This work was supported by the ICT R\&D program of MSIP/IITP (No. 2014-0-00077, Development of Global Multi-target Tracking and Event Prediction Techniques based on Real-time Large-scale Video Analysis) and by Institute of Information \& communications Technology Planning \& Evaluation (IITP) grant funded by the Korea government (MSIT) (No. 2021-0-00261, Development of High-Efficiency Programmable NPU and AI Camera Module for Smart Camera).
\end{acks}

\onecolumn

\begin{multicols}{2}
\bibliographystyle{ACM-Reference-Format}
\bibliography{camera-ready}
\end{multicols}

\newpage
\appendix
\section{Supplementary}
\subsection{Experimental Setting on Toy Example}
\label{sup:toy_example}
We generated four-classes toy datasets from \textit{Scikit-Learn}~\footnote{https://scikit-learn.org/stable/datasets/toy\_dataset.html} toolbox. 
Each class consists of 1000 training samples in 2D domain. The constructed model we constructed is stacked with a sequence of three consecutive ReLU-activated dense layers with 1000 hidden neurons each. We have also added a Dropout function with the probability of 50\%.

For the toy experiment, the model is constructed as follows,
\begin{itemize}
    \item Dense~(1000) $\rightarrow$ ReLU $\rightarrow$ Dropout~(0.5) $\rightarrow$ Dense~(1000) $\rightarrow$ ReLU $\rightarrow$ Dense~(1000) $\rightarrow$ ReLU $\rightarrow$ Dense~(k)
\end{itemize}
The model which takes 2D inputs will take an output that is the same number of classes (i.e. \textit{= 4}). We then sample a number of new data points (unseen during training) for the computation of the pruning criteria. For pruning, we remove a fixed number of 1000 of 3000 hidden neurons with the least relevance for prediction according to each criterion. This is equivalent to removing 1000 learned filters from the model. After pruning, we observed the changes in the decision boundary area and re-evaluated classification accuracy on the original 4000 training samples with the pruned model. Please note that after pruning, we directly show the decision boundary and accuracy as it is without fine-tuning step.

\subsection{Ablation Study}
\label{sec:Ablation study}

\subsubsection*{\textbf{Results in Data Quality}}
First, we see if our proposed method can achieve reasonable performances regardless of data quality. These results demonstrate that the performance of nuclear-norm based pruning is stable and independent of the data quality. Among the first 10 batches, we select a single batch of samples with 1) the lowest loss (called ``easy'' samples) and 2) the highest loss (called ``hard'' samples). In the previous pruning or neural architecture search (NAS) literatures, they use a small proxy dataset for searching and pruning the models, which means that it also gives a great impact with respect to pruning efficiency~\cite{DANAS2020}.

\begin{figure}[b!]
	\begin{center}
		\centerline{\includegraphics[width=0.6\linewidth]{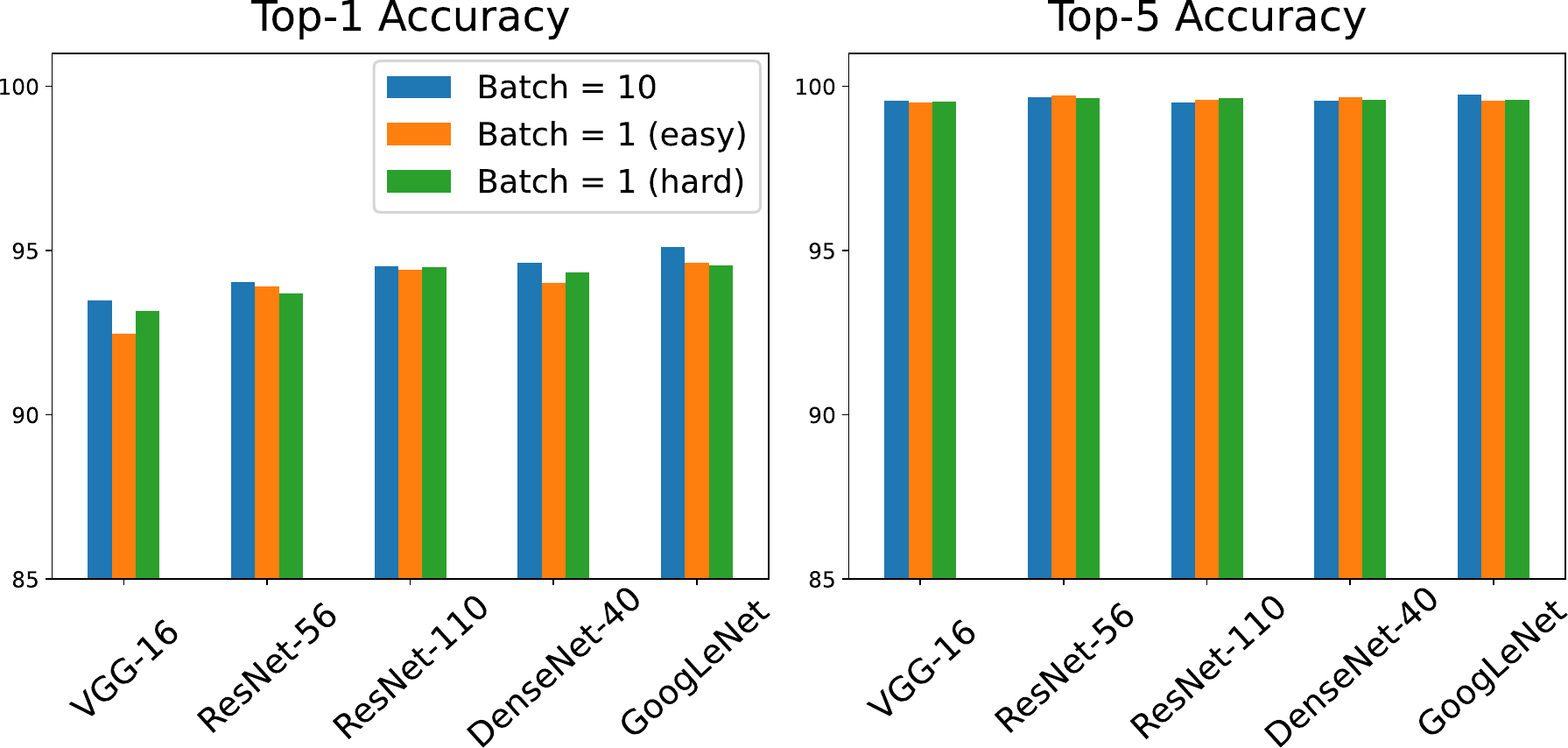}}
		\caption{Comparison study of Top-1 and Top-5 accuracies with 1) small (=batch of 10), 2) easy (=batch of 1), 3) hard (=batch of 1) dataset with five different neural network architectures.}
		\label{plot:result_comparison}
	\end{center}
\end{figure}

Figure~\ref{plot:result_comparison} shows comparison results of the Top-1 and Top-5 accuracy across small-batch (= 10), easy (= 1) and hard (= 1) samples on five different network architectures. We can observe that by using only a batch with easy as well as hard samples, our first ablation study found no significant differences across three different conditions (i.e. small-batch vs. easy vs. hard).
This experiment result demonstrates that competitive performance can be produced by NN-based filter pruning regardless without considering data quality for efficient pruning.

\subsubsection*{\textbf{Results in Data Quantity}}
\label{sup:result_quantuty}

From the practical point of view, compared to ImageNet, PASCAL VOC~\cite{PASCALVOC}, and COCO~\cite{COCO}, most of the private datasets have a smaller amount of data quantity which might be not guaranteed to be optimal for efficient pruning. In this manner, one of the interesting points in the pruning community is to see how large the amount of dataset we need for the proper pruning in terms of data quantity. Therefore, to evaluate the stability of the proposed criterion by data quantity, we perform a statistical test on 4 convolutional layers at regular intervals, called Kendall tau distance, to measure the pairwise similarity of two filter ranking lists of neighbor batches based on nuclear-norm to see the evolutionary change in increasing batch size. The equation for Kendall tau distance can be expressed as follows:

\begin{align}
    \label{eq:kendall1}
    K(\tau^1, \tau^2) = \frac{1}{n \times (n-1)} \sum_{(j,s), j \neq s} K_{js}^{\ast} (\tau^1, \tau^2)
\end{align} 
where $K_{js}^{\ast} (\tau^1, \tau^2)$ is assigned to 0 if $x_j$, $x_s$ are in the same order in $\tau^1$ and $\tau^2$ and 1 otherwise.

\begin{figure*}[t!]
	\begin{center}
		\centerline{\includegraphics[width=0.8\linewidth]{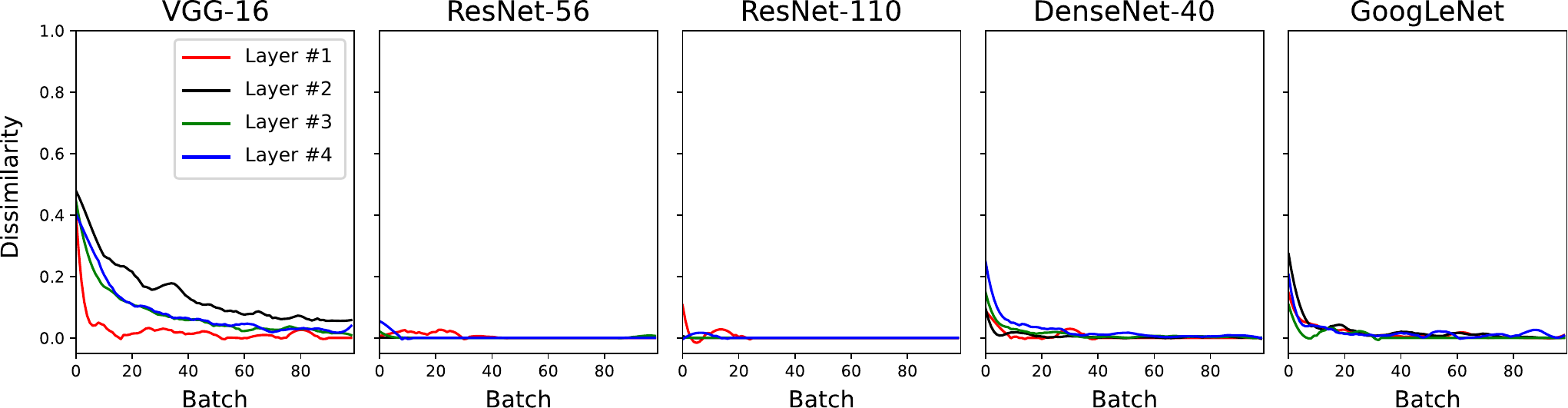}}
		\caption{Results of Kendall tau distance between filter ranking lists of two neighbour batch sizes. Here, values with y-axis is close to 0 when paired observations between two neighbour batches have a similar rank order and vice versa.}
		\label{plot:correlation}
	\end{center}
\end{figure*}

We empirically observe that the ranking order generated by the proposed criterion is stable and independent of the data quantity. Figure~\ref{plot:correlation} shows the similarity between neighbor of batches with Kendall tau distance. Here, we can observe that for ResNet-56/110, DenseNet-40, and GoogLeNet, there is a very close similarity of ranking order before the batch of ten which means the proposed method extracts stable ranking order indices layer-wisely, whereas VGG-16 observes the higher similarity between neighbor of batches after batch indices of 50 which indicates that it needs more data to get the stable ranking order.

\subsubsection*{\textbf{Results in Various Datasets and Various Deep Learning Modules}}
To check the further feasibility of the proposed method on the various module and the datasets, we perform additional experiments on the depthwise separable module on the CIFAR-100 and Imagewoof. The proposed method could compress 80.86\% / 88.51\% of the FLOPs~/parameters while it only shows 0.53\% of the performance drop on the  CIFAR-100 datasets for the MobileNetV1. Furthermore, for the MobileNetV2 on the Imagewoof dataset, the proposed method could compress 18.61\% / 22.32\% of the FLOPs/parameters while it only shows 0.28\% of the performance drop.

\begin{table}[]
    \caption{Pruning results on MobileNetV1 (CIFAR-100) and MObileNetV2 (Imagewoof).}
    \scalebox{0.90}{
        \begin{tabular}{ccccc} 
            \specialrule{1pt}{1pt}{1pt}
            Dataset   & Network     & Top-1 Acc ($\downarrow$)      & FLOPs ($\downarrow$\%)          & Params ($\downarrow$\%)   \\ \hline
            CIFAR-100 & MobileNetV1 & 66.68 (-0.53) & 17.78 (80.86)  & 0.38 (88.51) \\
            Imagewoof & MobileNetV2 & 91.24 (-0.28) & 487.54 (18.61) & 1.74 (22.32)\\
            \specialrule{1pt}{1pt}{1pt}
        \end{tabular}   
    }
\end{table}

\end{document}